\documentclass[conference]{IEEEtran}
\IEEEoverridecommandlockouts
\usepackage{cite}
\usepackage{amsmath,amssymb,amsfonts}
\usepackage{algorithmic}
\usepackage{url}
\usepackage{graphicx}
\usepackage{textcomp}
\usepackage{xcolor}

\def\BibTeX{{\rm B\kern-.05em{\sc i\kern-.025em b}\kern-.08em
    T\kern-.1667em\lower.7ex\hbox{E}\kern-.125emX}}

\makeatletter
 \let\old@ps@headings\ps@headings
 \let\old@ps@IEEEtitlepagestyle\ps@IEEEtitlepagestyle
 \def\confheader#1{%

 \def\ps@IEEEtitlepagestyle{%
 \old@ps@IEEEtitlepagestyle%
 \def\@oddhead{\strut\hfill#1\hfill\strut}%
 \def\@evenhead{\strut\hfill#1\hfill\strut}%
 }%
 \ps@headings%
 }
 \makeatother

\confheader{%
  \parbox{\textwidth}{%
    \hspace{0cm}2024 27th International Conference on Computer and Information Technology (ICCIT)\\
    \hspace{-6cm}20-22 December 2024, Cox’s Bazar, Bangladesh
  }
}

 \usepackage[pscoord]{eso-pic}
\newcommand{\placetextbox}[3]{
 \setbox0=\hbox{#3}
 \AddToShipoutPictureFG*{ \put(\LenToUnit{#1\paperwidth},\LenToUnit{#2\paperheight}){\vtop{{\null}\makebox[0pt][c]{#3}}}
 }
 }
 \placetextbox{.21}{0.055}{\small{979-8-3315-1909-4/24/\$31.00~\copyright 2024 IEEE}} 

\begin{document}

\title{Deep Learning for Breast Cancer Detection: Comparative Analysis of ConvNeXT and EfficientNet}

\author{\IEEEauthorblockN{Mahmudul Hasan}
\IEEEauthorblockA{\textit{Department of Computer Science and Engineering} \\
\textit{BRAC University, Dhaka, Bangladesh}\\
mahmudul.hasan5@g.bracu.ac.bd}
}

\maketitle

\begin{abstract}
Breast cancer is the most commonly occurring cancer worldwide. This cancer caused 670,000 deaths globally in 2022, as reported by the WHO. Yet since health officials began routine mammography screening in age groups deemed at risk in the 1980s, breast cancer mortality has decreased by 40\% in high-income nations. Every day, a greater and greater number of people are receiving a breast cancer diagnosis. Reducing cancer-related deaths requires early detection and treatment. This paper compares two convolutional neural networks called  ConvNeXT and EfficientNet to predict the likelihood of cancer in mammograms from screening exams. Preprocessing of the images, classification, and performance evaluation are main parts of the whole procedure. Several evaluation metrics were used to compare and evaluate the performance of the models. The result shows that ConvNeXT generates better results with a 94.33\% AUC score, 93.36\% accuracy, and 95.13\% F-score compared to EfficientNet with a 92.34\% AUC score, 91.47\% accuracy, and 93.06\% F-score on RSNA screening mammography breast cancer dataset.
\end{abstract}


\section{Introduction}
Breast Cancer (BC) is a kind of tissue cancer that generally affects the ducts and the inner layer of the milk glands or lobules [1]. Age [2], high hormone levels [3], race, economic status, and dietary iodine deficiency ([4], [5]) are the major risk factors for cancer. One stage of the multi-stage pathogenic process that leads to BC involves viruses [6]. With over a million new cases annually [7], it is the most common form of malignant neoplasms in women [8], and it is continuously increasing as both incidence and survival continue to climb in many parts of the world. This disease is the second most common cause of cancer-related death and the main cause of death for women between the ages of 45 and 55 [9]. BC is a complex disease that results from the interplay of hereditary and environmental risk factors [10]. Approximately 20\% of all cases of BC are familial in origin and are caused by a specific gene that predisposes to the disease [11]. The early phases of the disease are characterized by poorly presented symptoms, which postpone diagnosis. Being able to offer complete diagnosis and treatment services is a problem when managing BC. The percentage of people who can survive BC may rise if the disease can be stopped by early detection. 

One of the imaging techniques that is often used for early cancer screening is mammography, where an aberration can be classified as normal, benign, or malignant [12]. Two angles of each breast are imaged during an X-ray screening for mammograms. Human readers, skilled radiologists, assess and examine screening mammograms [13]. Double reading has been used in several nations because it was discovered to enhance evaluation results [14]. The diagnostic performance can be further enhanced by multiple readings—up to ten readers or more [15]. Beyond the double reading, there is still much space for improvement in the appraisal of mammograms. Several investigations have been carried out to assess tumor detection in mammography pictures. One study using the IRMA dataset used the VGG16 and ResNet50 models to distinguish between normal and abnormal tumors and got 94\% accuracy with VGG16 and 91.7\% with ResNet50 [16].  

Relevance Vector Machine [17] produced a 97\% accuracy rate for BC detection using the Wisconsin dataset. In contrast, [18] assigned separate weights to separate attributes depending on how well they could predict outcomes and produced 92\% accuracy using the weighted naive Bayes method. [19] developed a hybrid classifier in WEKA using decision trees and support vector machines, and the result was a 91\% accuracy rate. [20] employed a 93\% accuracy rate in feature selection using Linear Discriminant Analysis and trained the dataset using the Mamdani Fuzzy Inference Model, one of the fuzzy inference techniques. A study conducted using RF, k-NN and NB using the Wisconsin dataset showed 92.90\%, 94.20\% and 87.09\%, respectively, in terms of F1 score [20]. 

While advancements in imaging technologies, such as mammography, have significantly improved early detection, existing studies have not fully explored the potential of modern deep learning architectures for mammogram analysis. Moreover, many studies rely on older and smaller datasets or employ models that may lack the ability to generalize well to large, diverse datasets. The proportion of people who can survive breast cancer may rise if the disease can be stopped by early detection. In this paper, I address this gap by comparing two state-of-the-art convolutional neural networks, ConvNeXT-small and EfficientNetV2-S, for the classification of mammograms by performing comprehensive preprocessing to enhance the quality of mammogram images for model training. Unlike previous works that focus on traditional models or limited datasets, I utilize the RSNA screening mammography dataset to assess the efficacy of these architectures in distinguishing between normal and abnormal mammograms. Through this approach, I aim to demonstrate the potential of deep learning models in advancing the accuracy and reliability of breast cancer detection.

The remainder of this paper is structured as follows: Section II details the selected dataset. Section III describes the methodology, data preprocessing steps, model architecture, and training procedures. Section IV highlights the evaluation measures and experimental results, and Section V concludes the paper. 

\section{Dataset}

Radiographic breast pictures of female individuals from the RSNA Screening Mammography BC dataset are used in this study. The dataset is publicly available on Kaggle. There are approximately 20000 patients in the dataset, and each patient typically has two left breast photos and two right breast images, as shown in Figure 1. 

\begin{figure}[htbp]
\centerline{\includegraphics[width=0.5\textwidth]{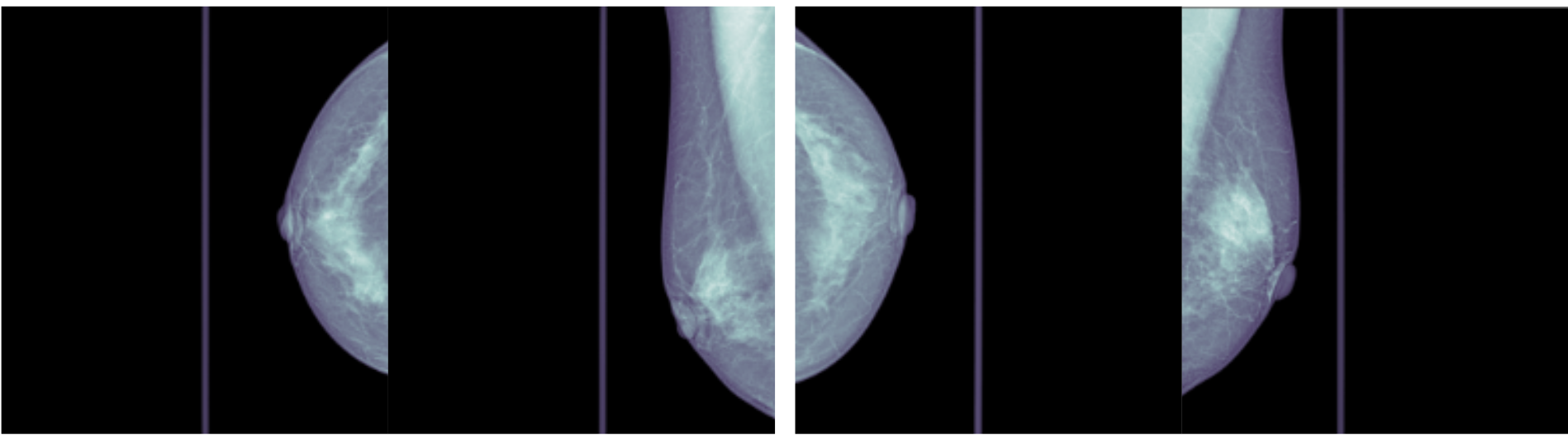}}
\caption{An illustration of a Mammogram from RSNA Dataset}
\label{fig1}
\end{figure}

The dataset exhibits several noteworthy patterns and characteristics, particularly relevant for understanding patient demographics and image-based cancer detection challenges. The age distribution of the patients, as illustrated in Fig. 2(a), reveals that most are older than 40, with two prominent peaks around the ages of 50 and 70. However, patient counts drop noticeably after 70, suggesting demographic or healthcare access factors at play. Cancer diagnoses are more likely in patients above 50, emphasizing the age-related nature of the disease. Despite these trends, the dataset poses an imbalanced classification problem, as the number of images showing cancer is exceedingly low, even fewer than the proportion of patients with cancer as shown in Fig.2(b). Regarding imaging data depicted in Fig.2(c), most patients have four images—two views per breast captured for both the left and right sides. In a few cases, more than 10 images per patient are available. 

Additionally, all cancer-diagnosed patients underwent biopsies, whereas only about 3\% of images from non-cancer cases resulted in follow-up biopsies, as shown in Fig.2(d). This stark difference highlights the potential value of the biopsy feature, especially when analyzed at the patient level. These insights collectively underline the complexity of the dataset and the need for careful feature engineering and validation strategies to address imbalances and improve cancer detection accuracy. 

\begin{figure}[htbp]
\centerline{\includegraphics[width=0.5\textwidth]{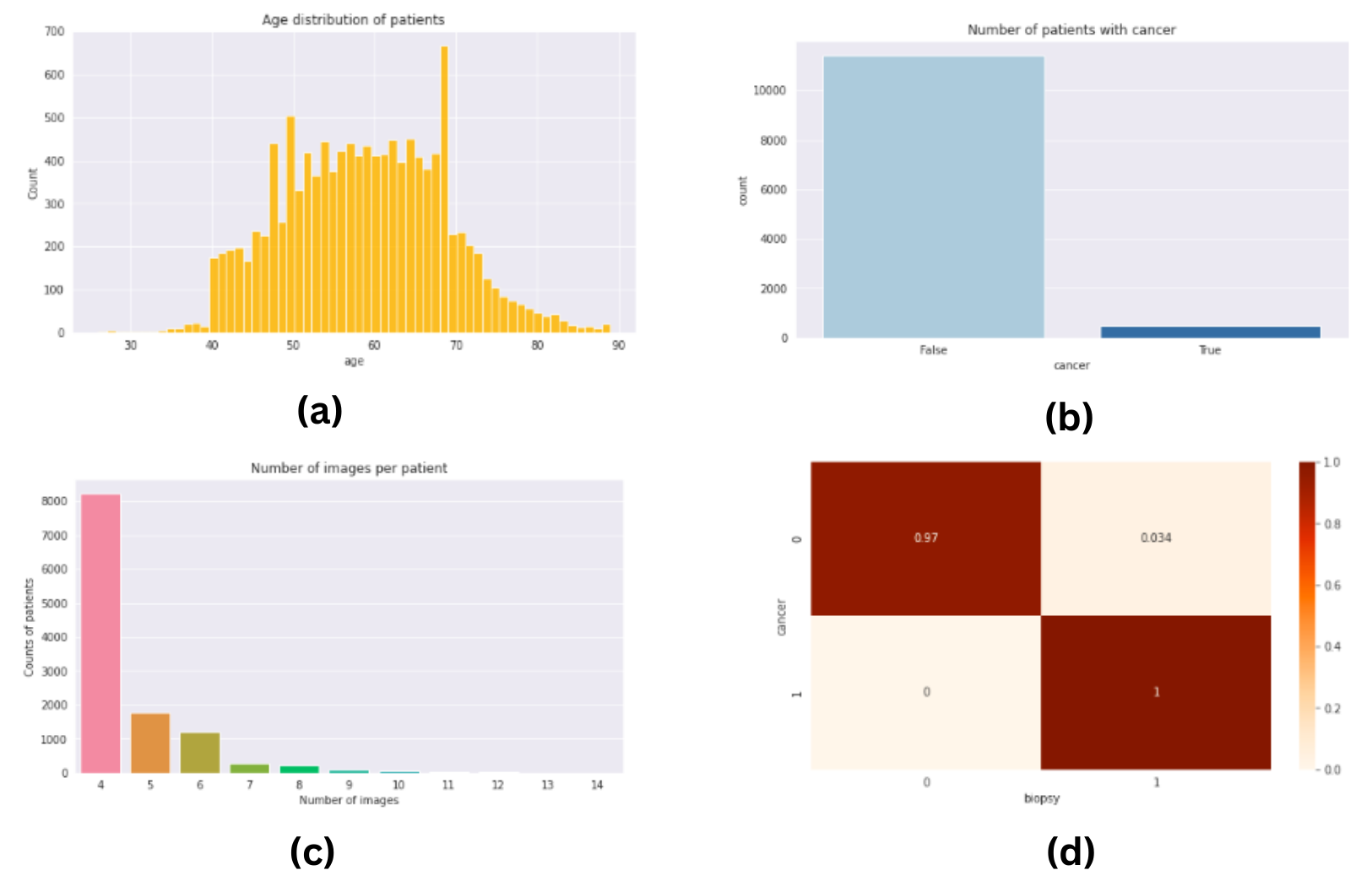}}
\caption{Cancer statistics: (a) Age distribution of patients; (b) Number of patients with cancer; (c) Number of images per patient; (d) Heatmap of biopsy feature.}
\label{fig2}
\end{figure}

\section{Methodology}
The overall structure of the entire BC prediction system is shown in Fig.3. To fit the network system, raw images from the RSNA Screening Mammography BC dataset are preprocessed using image resizing and transformation. Next, two CNN based model networks: ConvNeXT-small and EfficientNetV2-S are used for identification of BC in screening mammography. The goal is to identify cases of BC in mammograms from screening exams. Performance evaluation indicators like AUC, Accuracy, and F-score are used to quantify the capability of classification models.

\begin{figure}[htbp]
\centerline{\includegraphics[width=0.38\textwidth]{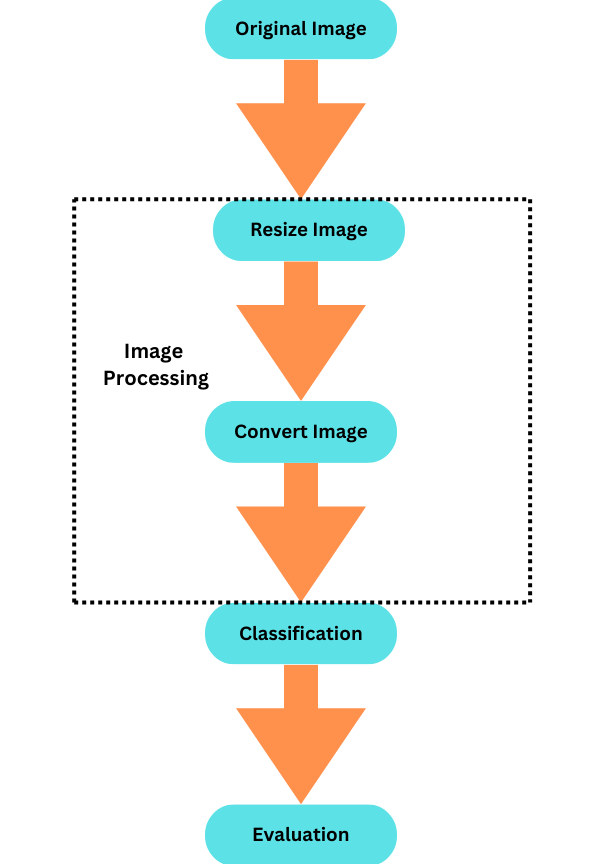}}
\caption{Top Level Overview of the Breast Cancer Prediction System}
\label{fig3}
\end{figure}

\subsection{Image Preprocessing}
\begin{enumerate}
    \item Image Cropping: To preserve texture and detail within a fixed resolution, ROI (Region of Interest) cropping was employed. The YOLOX-nano detector (configured at 416x416) was utilized for this purpose. Unlike traditional rule-based methods, deep learning-based detectors offer several advantages, including generating smaller resulting images, achieving better aspect ratio compatibility, and focusing more effectively on the breast region.
    \item Apply Windowing: Windowing is applied to the cropped ROI to refine the pixel intensity distribution, enhancing the visibility of important features. This process adjusts the intensity values within a specified range, often defined by a window level and window width.
    \item Rescaling and Padding: To ensure uniformity and compatibility with the input requirements of downstream models, the cropped ROI undergoes isotropic rescaling followed by padding. Isotropic rescaling involves resizing the ROI while preserving its aspect ratio, thus avoiding geometric distortions. The resized image is then padded with constant pixel values to achieve a target size. 
\end{enumerate}
% \subsubsection{Image Cropping}
% To preserve texture and detail within a fixed resolution, ROI (Region of Interest) cropping was employed. The YOLOX-nano detector (configured at 416x416) was utilized for this purpose. Unlike traditional rule-based methods, deep learning-based detectors offer several advantages, including generating smaller resulting images, achieving better aspect ratio compatibility, and focusing more effectively on the breast region.
% \subsubsection{Apply Windowing}
% Windowing is applied to the cropped ROI to refine the pixel intensity distribution, enhancing the visibility of important features. This process adjusts the intensity values within a specified range, often defined by a window level and window width.
% \subsubsection{Isotropic Rescale and Padding}
% To ensure uniformity and compatibility with the input requirements of downstream models, the cropped ROI undergoes isotropic rescaling followed by padding. Isotropic rescaling involves resizing the ROI while preserving its aspect ratio, thus avoiding geometric distortions. The resized image is then padded with constant pixel values to achieve a target size. 

\subsection{Classification}
Convolutional neural network (CNN) is regularly used in various projects to work with image data. CNN uses 3-dimensional data to perform tasks like object recognition and image categorization. Their three main layer types are convolutional layer, pooling layer, and fully connected layer. The CNN becomes more complex and has a greater range of picture recognition with each layer. Previous layers draw attention to basic elements like boundaries and colors. The CNN begins to recognize the object's main components or forms as the image input moves through its layers, eventually identifying the desired object. Neurons in convolutional layers will only connect to a tiny part of the layer above. By doing this, the computational complexity is decreased and CNN is able to fully utilize the input data's 2D structure. CNN can therefore regularly generate better outcomes in image recognition when compared to other deep learning architectures [21]. 
% \begin{figure}[htbp]
% \centerline{\includegraphics[width=0.5\textwidth]{images/cnn.png}}
% \caption{Architecture of CNN}
% \label{fig4}
% \end{figure}
\subsection{Models of CNN}
\subsubsection{ConvNeXT}
The ConvNeXT model is a pure convolutional network (ConvNet) that was inspired by the architecture of Vision Transformers. It is stated in ``A ConvNet for the 2020s" [22] to outperform them. In the ConvNeXT architecture, convolutional layers are the foundational building blocks, followed by fully connected layers for final predictions. The convolutional layers play a pivotal role in feature extraction by capturing spatial hierarchies in the input data, making these models exceptionally adept at processing visual datasets like images and videos. By leveraging these capabilities, ConvNeXT has proven to be highly effective for tasks requiring detailed visual analysis, including classification, segmentation, and object detection. The architecture of ConvNeXT is displayed in Fig.4.
\begin{figure}[htbp]
\centerline{\includegraphics[width=0.3\textwidth]{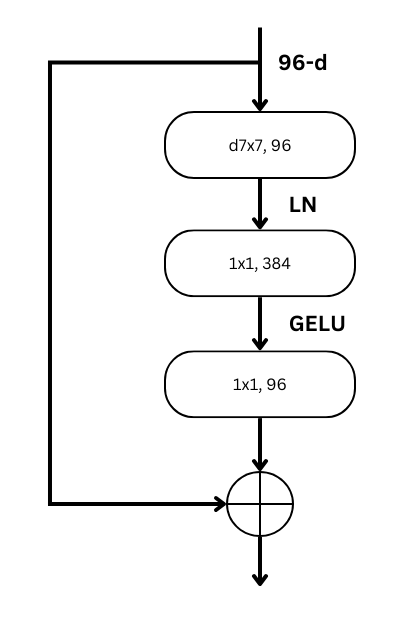}}
\caption{Architecture of ConvNet}
\label{fig4}
\end{figure}
\subsubsection{EfficientNet}
Convolutional neural networks (CNNs) in the EfficientNet family are designed to operate at high performance while requiring less computing power than earlier architectures. Mingxing Tan and Quoc V. Le of Google Research first presented it in their paper published in 2019 [23]. EfficientNet is based on a ground-breaking scaling mechanism which utilizes a compound coefficient to scale depth, width, and resolution equally. Numerous of EffNets models (B0 to B3 and S, M, L) are more compact than alternative topologies with fewer parameters but greater efficiency. The efficientNet architecture as depicted in Fig.5.
\begin{figure}[htbp]
\centerline{\includegraphics[width=0.5\textwidth]{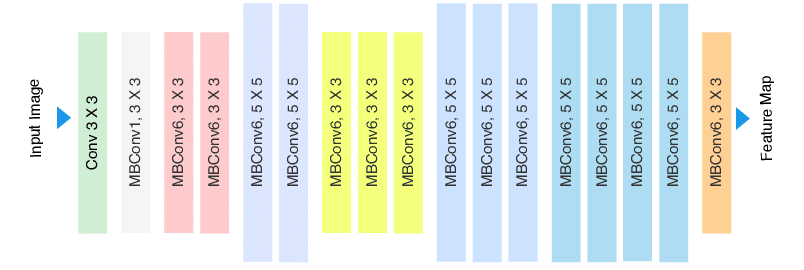}}
\caption{Architecture of EfficientNet}
\label{fig5}
\end{figure}

\subsection{Training}
To train the models effectively, a four-fold cross-validation strategy was implemented, resulting in four models corresponding to the four data splits. The training process evolved through multiple stages to address unexpected results observed during the initial attempts, leading to refinements in configuration and strategy. The final solution's training used multiple external datasets, including VinDr-Mammo, MiniDDSM, CMMD, CDD-CESM, and BMCD. \\

The models were configured with a set of carefully chosen hyperparameters and design choices to ensure effective training and robust performance. The loss function employed was Binary Cross-Entropy (BCE) without class weighting, ensuring a straightforward optimization process. To address class imbalance, a sampler was used to upsample positive samples per epoch, with each batch containing at least one positive sample. A batch size of 8 was selected to balance memory constraints and computational efficiency. Training optimization was carried out using Stochastic Gradient Descent (SGD) with momentum, paired with a cosine learning rate scheduler. Regularization techniques included a dropout rate and a drop path rate, preventing overfitting. Max pooling was employed as the global pooling method for feature aggregation. Soft positive label values were selected to refine the training process, improve convergence, and enhance class separation. This configuration, alongside iterative refinements, ensured robust model training across various dataset splits.

\section{RESULT AND DISCUSSION}

\subsection{Evaluation Metrics}
Several evaluation metrics were employed to evaluate the effectiveness of the models.
\begin{itemize}
    \item Precision: Out of all cases projected as positive (TP+FP), precision quantifies the percentage of accurately predicted positive cases (TP). It emphasizes how accurate the optimistic forecasts were.
    \begin{equation}
    Precision=\frac{TP}{TP \ + \ FP} 
    \end{equation}
    \item Recall: The percentage of accurately anticipated positive cases (TP) among all actual positive instances (TP+FN) is known as recall. It shows how well real positives are captured by the model.
    \begin{equation}
    Recall=\frac{TP}{TP \ + \ FN} 
    \end{equation}
    \item Accuracy: The percentage of accurately predicted cases (TP+TN) out of all cases in the dataset (TP+TN+FP+FN) is known as accuracy.
    \begin{equation}
    Accuracy=\frac{TP+TN}{TP+TN+FP+FN} 
    \end{equation}
    \item F1-score: The harmonic mean of recall and precision is known as the F1-Score. It is helpful for assessing models with an unbalanced class distribution since it strikes a balance between the two metrics.
    \begin{equation}
    F1-score=\frac{2*Precision*Recall}{Precision + Recall} 
    \end{equation}
    \item AUC: The performance of the classifier over all potential classification thresholds is summed up by a single figure called the ROC AUC score. The classifier's ability to discern between positive and negative classes is calculated by the score. 
    
\end{itemize}

\subsection{Results}
The model’s output reflects the probability of cancer in the relevant column, and its effectiveness was evaluated using standard performance metrics such as AUC, accuracy, and F1 score, summarized in Table 1. These metrics contribute a thorough view of the model's performance, balancing both classification accuracy and the ability to handle imbalanced data. The ConvNeXT-small model exhibited superior performance on the RSNA BC dataset, achieving an AUC of 0.9433, accuracy of 0.9336, precision of 0.9321, recall of 0.9524 and F1 score of 0.9513, outperforming the EfficientNetV2-S model, which achieved an AUC of 0.9234, accuracy of 0.9147, precision of 0.9244, recall of 0.9305 and F1 score of 0.9306. This highlights the robustness and generalization capability of ConvNeXT-small, particularly in extracting meaningful features relevant to cancer prediction.
 
\begin{table}[htbp]
\caption{Comparison of the overall effectiveness of EfficinetNetV2-S and Convnext-small}
\begin{center}
\begin{tabular}{|c|c|c|c|c|c|}
\hline
 Model & AUC & Precision & Recall & Accuracy & F-score \\
 \hline
 ConvNeXT-S & 0.9433 & 0.9321 & 0.9524 & 0.9336 & 0.9513 \\
 \hline
 EffNetV2-S & 0.9234 & 0.9244 & 0.9305 & 0.9147 & 0.9306 \\
 \hline
\end{tabular}
\end{center}
\end{table}

Figure 6 illustrates a comparison between the proposed models and the feature extraction and concatenation approach by Jafari and Karami (2022) [24]. Their concatenation-based model, coupled with a neural network classifier, achieved a peak accuracy of 92\%, AUC of 96\%, and F1 score of 94\% on the RSNA dataset. Although this method shows competitive AUC, the ConvNeXT-small model slightly surpasses it in overall performance metrics.

\begin{figure}[htbp]
\centerline{\includegraphics[width=0.5\textwidth]{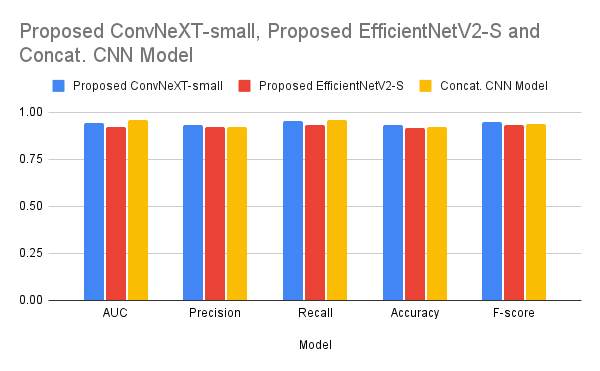}}
\caption{Comparison of Proposed Models with Concat. Model [24]}
\label{fig6}
\end{figure}

In Figure 7, the proposed models are compared with four CNN architectures evaluated by Huynh et al. (2023) [25]: VGG, GoogLeNet, EfficientNet, and Residual Networks. Among these, the fine-tuned EfficientNet model achieved the highest accuracy of 89\% and an AUC of 92\%. This comparison underscores ConvNeXT’s capacity to adapt to complex datasets like RSNA and outperform traditional CNNs.

\begin{figure}[htbp]
\centerline{\includegraphics[width=0.5\textwidth]{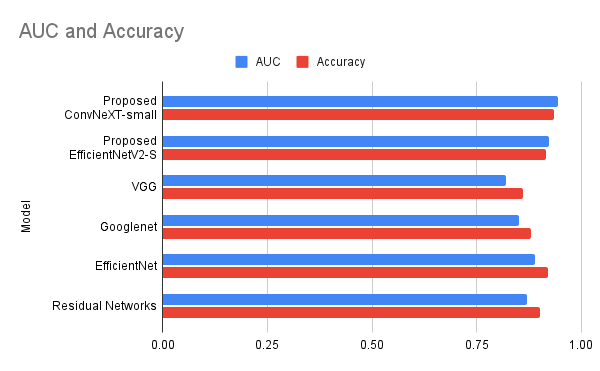}}
\caption{Comparison of Proposed Models with respect to AUC and Accuracy}
\label{fig7}
\end{figure}

\subsection{Discussion}
In this study, I developed and evaluated two deep CNN models for BC detection using the RSNA mammography dataset. Among the models, ConvNeXT-Small outperformed EfficientNetV2-S across widely recognized evaluation metrics such as AUC, accuracy, and F-score. The proposed models were also benchmarked against existing methods applied to the same dataset, with ConvNeXT-Small demonstrating superior performance.

These results highlight the critical role of selecting an appropriate model architecture for medical imaging tasks. ConvNeXT’s superior performance can be attributed to its architectural design, which incorporates transformer-inspired features to enhance the capabilities of convolutional networks. The findings underscore that deep CNNs can provide highly accurate and reliable BC detection. Incorporating multi-modal data, such as clinical records and genetic information, alongside mammography images could further enhance diagnostic accuracy and robustness.

\section{CONCLUSION}
This paper compares the performance of ConvNeXT and EfficientNet deep learning techniques for the detection of BC in mammograms obtained from screening tests. The results show that the ConvNext model achieved a greater accuracy of 93.36\% on the test set, outperforming the EfficinetNet model. The findings also showed the promise of deep learning techniques for the diagnosis and screening of BC as well as the importance of using extensive and varied datasets for model evaluation and training. By accurately and promptly detecting malignant tumors, the suggested methods can help lower the death and morbidity rates associated with breast cancer.

% \bibliographystyle{plain}
% \bibliography{reference.bib}

\end{document}